\pdfoutput=1

\documentclass[11pt]{article}

\usepackage[]{acl}
\usepackage{times}
\usepackage{latexsym}
\usepackage{subfig}
\usepackage{multirow}
\usepackage{url}
\usepackage{amsfonts}
\usepackage{amssymb}
\usepackage{mathtools}
\usepackage{amsthm}
\usepackage[T1]{fontenc}

\usepackage[utf8]{inputenc}

\usepackage{microtype}

\title{Investigating Chain-of-thought with ChatGPT for Stance Detection on Social Media}

\author{Bowen Zhang\textsuperscript{1}, 
Xianghua Fu\textsuperscript{1}, 
Daijun Ding\textsuperscript{1}, 
Hu Huang\textsuperscript{2}, 
Genan Dai\textsuperscript{1}, 
Nan Yin\textsuperscript{3},
Yangyang Li\textsuperscript{4\ *}, Liwen Jing\textsuperscript{5}  \thanks{\ \ Corresponding authors: Yangyang Li and Liwen Jing (email: liyangyang@live.com and ljing@x-institute.edu.cn)}\\
  \textsuperscript{1}College of Big Data and Internet, Shenzhen Technology University, Shenzhen, China\\
    \textsuperscript{2}School of Cyberspace Science and Technology, \\University of Science and Technology of China, Hefei, China\\
    \textsuperscript{3} Mohamed bin Zayed University of Artificial Intelligence\\
    \textsuperscript{4}Academy of Cyber, Beijing, China\\
  \textsuperscript{5} Shenzhen X-Institute, Shenzhen, China\\
  }

\begin{document}
\maketitle
\begin{abstract}
Stance detection predicts attitudes towards targets in texts and has gained attention with the rise of social media. Traditional approaches include conventional machine learning, early deep neural networks, and pre-trained fine-tuning models. However, with the evolution of very large pre-trained language models (VLPLMs) like ChatGPT (GPT-3.5), traditional methods face deployment challenges. The parameter-free Chain-of-Thought (CoT) approach, not requiring backpropagation training, has emerged as a promising alternative. This paper examines CoT's effectiveness in stance detection tasks, demonstrating its superior accuracy and discussing associated challenges.
\end{abstract}

\section{Introduction}
Stance detection on social media is an important topic in research communities of both natural language processing (NLP) and social computing \cite{kuccuk2020stance, li2023stance}. 
The goal of stance detection is to automatically predict the attitude (i.e., \textit{favor}, \textit{against}, or \textit{neutral}) of opinionated tweets (text) with a specified target. 

Early research works on stance detection mainly adopted rule-based and traditional machine learning techniques \cite{anand2011cats, walker2012stance}. 
For example, the effective algorithms for the classifiers are support vector machine (SVM), logistic regression, naive bayes, decision tree and etc \cite{aldayel2021stance}.
As deep learning methods evolved, deep neural networks (DNNs) rapidly became the prevailing techniques for stance detection. These methods employ neural networks with varying structures and connections to develop the desired stance classifier, which can be classified into conventional DNN models, attention-based DNN models, and graph convolutional network (GCN) models.
Convolutional neural networks (CNNs) and long short-term memory (LSTM) models are among the most commonly used conventional DNN models \cite{augenstein2016stance, 
 jiang2019hierarchical}; attention-based methods primarily leverage target-specific information as the attention query and implement an attention mechanism for inferring stance polarity \cite{dey2018topical}; and GCN methods propose a graph convolutional network to model the relationship between the target and text \cite{zhang2020enhancing}.

\begin{figure*}[!htbp]
	\centering
	\includegraphics[width=0.8\textwidth]{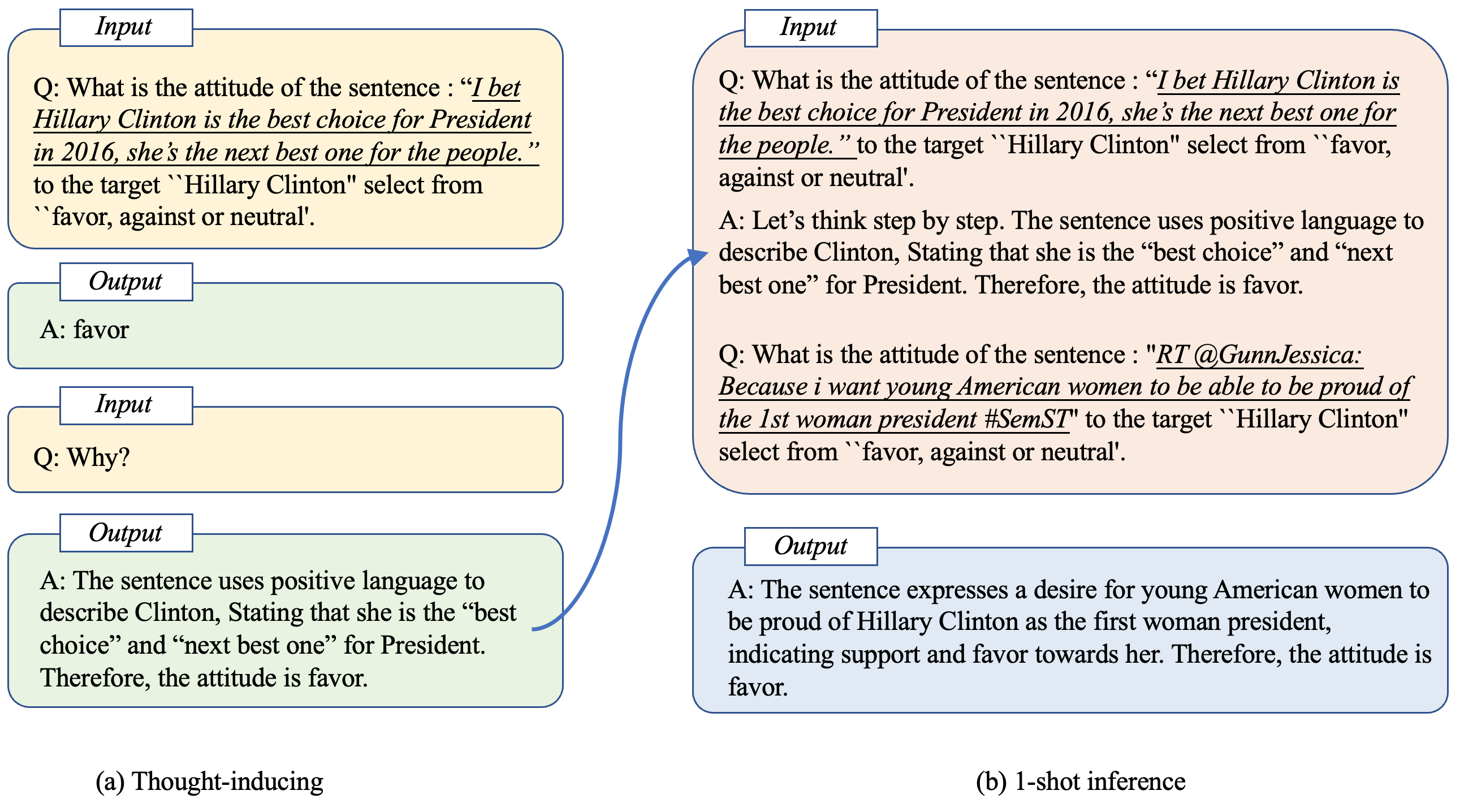}
	\caption{A demonstration of StSQA CoT prompting for stance detection. (The underlined text is the original text).}
	\label{Fig1}
\end{figure*}

Subsequently, with the great success of bidirectional encoder representations from transformers (BERT) model, a new NLP research paradigm emerges which is utilizing large pre-trained language models (PLM) with a fine-tuning (or prompt-tuning) process. 
Fine-tuning models adapt PLMs by building a stance classification head on top of the ``{$<$cls$>$}'' token, and fine-tune the whole model \cite{wei2022emergent}. 
This pre-train and fine tuning paradigm provides exceptional performance for most NLP downstream tasks including stance detection, because the abundance of training data enables PLMs to learn enough general-purpose features and knowledge for modeling different languages. 
The main idea of prompt-based methods is mimicking PLMs to design a template suitable for classification tasks and then build a mapping (called verbalizer) from the predicted token to the classification labels to perform class prediction, which bridges a projection between the vocabulary and the label space. The prompting strategies provide further improvements for stance detection performance\cite{shin2020autoprompt}. 

As the number of model parameters continues to grow, language model performance has experienced consistent advancements. The advent of very large pre-trained language models (VLPLMs), such as ChatGPT, which is built upon the GPT series architecture, has revolutionized the progress in various language-related tasks. These models have showcased remarkable efficacy across a multitude of tasks, outperforming existing pre-training and fine-tuning frameworks to achieve state-of-the-art results, even in zero-shot scenarios where no training data is available. Such progress marks a significant breakthrough, highlighting the immense potential for ongoing research and the application of large-scale models in the future.
However, contemporary VLPLMs have already attained a colossal scale, encompassing hundreds of billions of parameters. This renders the cost of pre-training and fine-tuning methods prohibitively expensive. Consequently, this article investigates how to effectively leverage VLPLMs for stance detection tasks, motivated by the growing need to address these challenges.

In this study, we investigate the efficacy of the chain-of-thought (CoT) prompting strategy when applied to ChatGPT (GPT-3.5) for stance detection tasks. The CoT approach involves utilizing templates as prompts to harness the model's capabilities more effectively \cite{dua2022successive, zhou2022least}. To the best of our knowledge, this constitutes the first research endeavor that employs the CoT technique within the context of stance detection tasks.
Our experimental results indicate that CoT methods can achieve state-of-the-art (SOTA) or comparable performance on widely used datasets, using a straightforward prompt without the need for training. Furthermore, we also identify and discuss certain limitations that may constrain the performance of CoT when utilizing VLPLMs.

\section{Methods and Results}
\textbf{Task definition.}
We use $X = \{x, p\}_{i=1}^{N}$ to denote the collection of data, where each $x$ denotes the input text and $p$ denotes the corresponding target. $N$ represents the number of instances.
Stance detection aims to predict a stance label for the input sentence $x$ towards the given target $p$ by using the stance predictor constructed by CoT. 

\textbf{Method.}
In this Section, we reveal the performance of the conventional CoT method for stance detection.
Specifically, we evaluate two distinct CoT methodologies, which are defined as direct question-answering (DQA) and step-by-step question-answering (StSQA), respectively.

(1) For DQA, we directly ask the ChatGPT model the stance polarity of a certain tweet towards a specific target.
For example, given the input: ``{RT \@GunnJessica: Because i want young American women to be able to be proud of the 1st woman president \#SemST}'',  the question for ChatGPT input is: {``What is the attitude of the sentence: "RT \@GunnJessica: Because i want young American women to be able to be proud of the 1st woman president \#SemST'' to the target ``Hillary Clinton" select from ``favor, against or neutral''.} For this particular example, ChatGPT returns a correct result. 

(2) StSQA prompting method teaches language models to solve the stance detection by providing a 1-shot example. 
This process comprises two stages:
1) Thought-inducing. The first stage is to construct the question-answer pair (QAP). We first feed the constructed question into the VLPLM and acquire the explanation of the reason for prediction. The example is given in Figure \ref{Fig1} (a). 
The sensitivity of QAP is discussed in Section III.
2) The second stage encompasses inferencing the tweet's stance using the provided QAP. In this context, the QA prompt serves as an example for CoT. An example is presented in Figure \ref{Fig1} (b).

\begin{table}[htbp]
\caption{Performance comparison on SemEval-2016 dataset and VAST dataset with zero shot setup.}
\begin{center}
\begin{tabular}{l|cccc|c}
\hline
\multirow{2}{*}{Model}    & \multicolumn{4}{c|}{SEM16 (F$_{avg}$)} & VAST  \\ 
    & HC   & FM   & LA & DT &  (F$_{m}$) \\ \hline
Bicond$^\dag$   & 32.7 & 40.6 & 34.4 & 30.5 & 41.0\\
CrossNet$^\dag$  & 38.3 & 41.7 & 38.5 &35.6 & 45.5 \\
SEKT     & 50.1 & 44.2 & 44.6 & 46.8 & 41.1 \\
TPDG$^\dag$    & 50.9 & 53.6 & 46.5 & 47.3& 51.9 \\
Bert\_Spc$^\dag$      & 49.6 & 41.9 & 44.8 & 40.1& 65.3 \\
Bert-GCN$^\dag$  & 50.0 & 44.3 & 44.2 & 42.3& 68.6\\
PT-HCL$^\dag$   & 54.5 & 54.6 & 50.9 & 50.1& \textbf{71.6}\\ \hline
DQA  &   {78.0}    &  \textbf{69.0}  &   {59.3} & {71.3} & 62.3 \\  
StSQA  &   \textbf{78.9}    &  {68.7}  &   \textbf{61.5}  & \textbf{71.6}& 68.9 \\ \hline 
\end{tabular}
\end{center}
\label{tab1}
\end{table}

\begin{table}[htbp]
\small
    \caption{Performance comparison on SemEval-2016 dataset with in-domain setup.}
\begin{center}
\begin{tabular}{lllllll}
\hline
Methods           & \multicolumn{2}{c}{FM}         &  \multicolumn{2}{c}{LA}        &    \multicolumn{2}{c}{HC}    \\ \cline{2-7}
                  & F$_m$ & F$_{avg}$ & F$_m$ & F$_{avg}$ & F$_m$ & F$_{avg}$   \\ \hline
BiLSTM            & 48.0   & 52.2     & 51.6   & 54.0     & 47.5   & 57.4        \\
BiCond           & 57.4   & 61.4     & 52.3   & 54.5     & 51.9   & 59.8         \\
TextCNN          & 55.7   & 61.4     & 58.8    & 63.2      & 52.4   & 58.5         \\
MemNet          & 51.1   & 57.8     & 58.9   & 61.0     & 52.3    & 60.3         \\
AOA               & 55.4   & 60.0     & 58.3   & 62.4     & 51.6   & 58.2        \\
TAN              & 55.8   & 58.3     & 63.7   & \textbf{65.7}     & 65.4   & 67.7        \\
ASGCN          & 56.2   & 58.5     & 59.5   & 62.9     & 62.2   & 64.3         \\
Bert\_Spc        & 57.3    & 60.6      & 64.0   & 66.3    & 65.8   & 69.1         \\
TPDG             & 67.3    & /        &     \textbf{74.7}  &  /       & 73.4   &  /             \\ \hline
DQA  & {68.4}   & \textbf{69.0}    & 58.2   & 59.3    & \textbf{79.5 }  & {78.0}     
\\
StSQA  & \textbf{68.7}   & {68.7}    & 61.8   & 61.5    & {78.9 }  & \textbf{78.9}     
\\				
\hline
\end{tabular}
\end{center}
    \label{tab2}
\end{table}

\begin{table}[htbp]
\small
    \centering
    \caption{Performance comparison on P-Stance dataset with in-domain setup.}
\begin{tabular}{lllllll}
\hline
Methods           & \multicolumn{2}{c}{Trump}         &  \multicolumn{2}{c}{Biden}        &    \multicolumn{2}{c}{Bernie}    \\ \cline{2-7}
                  & F$_m$ & F$_{avg}$ & F$_m$ & F$_{avg}$ & F$_m$ & F$_{avg}$   \\ \hline
BiLSTM               & 69.7  & 72.0     & 68.7  & 69.5       & 63.8      & 63.9   \\
BiCond             & 70.6  & 73.0      & 68.4  & 69.4     & 64.1   & 64.6    \\
TextCNN           & 76.9   & 77.2       & 78.0    & 78.2     & 69.8     & 70.2     \\
MemNet        & 76.8  & 77.7    & 77.2   & 77.6        & 71.4    & 72.8      \\
AOA        & 77.2    & 77.7   & 77.7   & 77.8    & 71.2      & 71.7     \\
TAN                 & 77.1  & 77.5    & 77.6   & 77.9    & 71.6       & 72.0      \\
ASGCN         & 76.8  & 77.0      & 78.2   & 78.4     & 70.6    & 70.8      \\
Bert\_Spc              & 81.4  & 81.6       & 81.5  & 81.7   & 78.3   & 78.4            \\ \hline
DQA    & {82.8}  & {83.2}   & {82.3}   & {82.0 }   & {79.4 }  & {79.4}     
\\
StSQA    & \textbf{85.4}  & \textbf{85.7}   & \textbf{82.4} & \textbf{82.8}  & \textbf{80.8} & \textbf{80.8}     
\\					
\hline
\end{tabular}
        \label{tab3}
\end{table}

\textbf{Datasets and Evaluation Matrix.}
We evaluated the CoT model using SemEval-2016 \cite{StanceSemEval2016}, VAST \cite{allaway2020zero}, and P-Stance datasets \cite{li2021p}. SemEval-2016 comprises 4870 annotated tweets for 6 targets, including Hillary Clinton, Feminist Movement, and Legalization of Abortion. VAST is a zero-shot stance detection dataset with 4003 training samples and separate dev and test sets. P-Stance contains 21,574 political tweets with stance annotations for ``Donald Trump'', ``Joe Biden'', and ``Bernie Sanders''. We assessed performance using F$_{avg}$ and macro-F1 score F${_m}$.

\textbf{Baselines}.
We adopt several stance detection methods as baselines.
{Bicond} \citep{augenstein2016stance}, {CrossNet} \citep{xu2018cross}, 
{SEKT} \citep{zhang2020enhancing}, {MemNet} \citep{TangQL16}, {AOA} \citep{huang2018aspect}, TAN \citep{du2017stance},
{ASGCN} \citep{zhang2019aspect},
{Bert\_spc} \citep{BERT},Bert-GCN \citep{linbertgcn} and PT-HCL \citep{liang2022zero}.

\textbf{Results.}
We constructed both zero-shot stance detection and in-domain stance detection setups for results comparison. 
Please note that in existing methods, the zero-shot learning setting refers to utilizing data from other targets for training and predicting in the unseen target domain. In contrast, DQA does not require any training samples, while StSQA requires one sample as a prompt example.
We also compared our zero-shot results of ChatGPT with other mainstream stance detection models in an in-domain setup, which means these models are optimized with 80\% tweets as training data. The 
results are summerized in Table \ref{tab1} to \ref{tab3}.
The results show that StSQA achieves SOTA results in zero-shot setup.
For example, StSQA achieves a 16.6\% improvement on average compared with the best competitor PT-HCL in zero-shot setup.
Compared with in-domain setup, where these methods first learned from 80\% training corpus, ChatGPT still yields better performance than all the baselines in most tasks.

\section{Discussion}
The findings presented in Section III highlight the emergent capabilities of ChatGPT (GPT-3.5) for stance detection tasks. Utilizing a straightforward prompt that directly queries the VLPLMs, without requiring any training, ChatGPT yields exceptional results.

The introduction of ChatGPT has the potential to significantly reshape the entire research field. Based on the existing data results and experimental process, we found three potential problems that could limit the performance of ChatGPT in stance detection tasks:

(1) \textbf{The stance biases of ChatGPT}.
In our experiments, we observed that ChatGPT exhibits potential biases towards specific targets, which we have identified as stance bias issues. For example, for the ``legalization of abortion" target (LA in SemEval-2016), we discovered that ChatGPT inherently supports this topic, consequently leading to an excessive number of neutral tweets being classified as supportive. 
As shown in Table II, ChatGPT's performance on the LA target is notably inferior compared to other in-domain models. This discrepancy can be attributed to ChatGPT exhibiting a specific stance bias towards the LA target.
This issue also arises in the VAST dataset.
Therefore, in future work, we can explore how to identify such underlying biases and, further, how to devise corresponding prompting strategies that enable large-scale models to mitigate these biases, thereby bolstering stance detection tasks.

\begin{figure}[htbp!]
	\centering
	\includegraphics[width=0.45\textwidth]{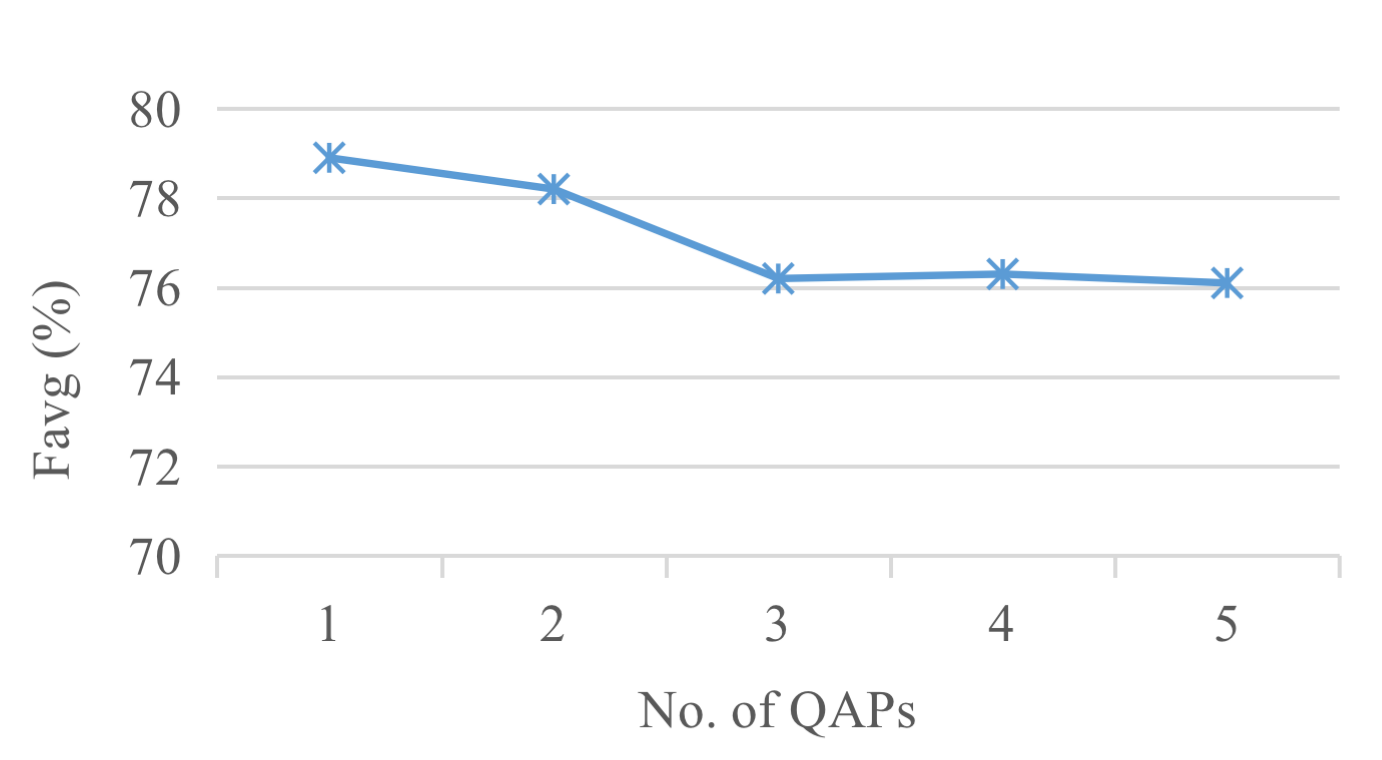}
	\caption{The experimental results for LA target with the number of QAPs.}
	\label{Fig122}
\end{figure}

\begin{figure}[htbp!]
	\centering
	\includegraphics[width=0.48\textwidth]{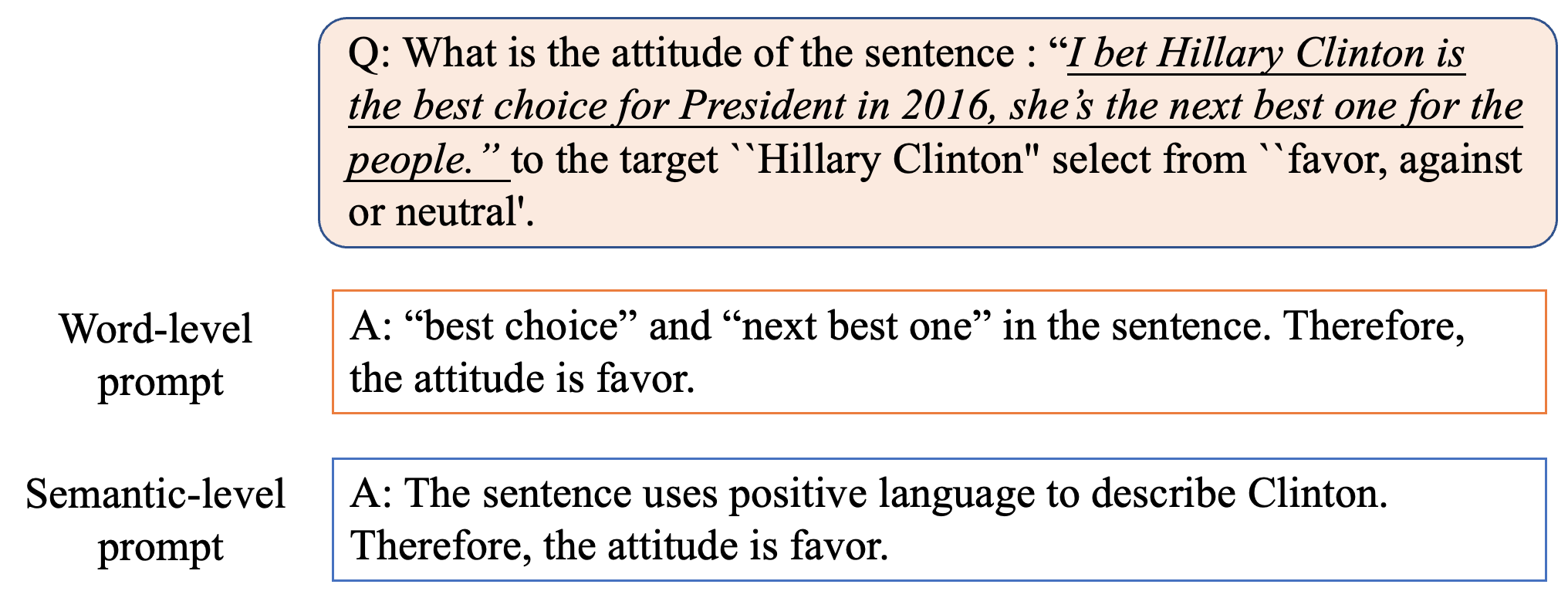}
	\caption{The difference between word-
level and semantic-level prompts.}
	\label{Fig222}
\end{figure}

(2) \textbf{Automatic selection of QAP}.
Second, we found that selecting appropriate QAPs as examples has a significant impact on the results. For instance, during our experiments, we discovered that multiple QAPs might confuse ChatGPT's reasoning abilities, thereby reducing its predictive performance. 
In this experiment, we test what the best number of QAPs should be on LA target. 
We run the experiments with the number of QAPs as one of (1, 2, 3, 4, 5).
The experimental results show in Figure \ref{Fig122}.

Besides, for single QAPs, a more specific CoT interpretation (influencing the stance from the word level) can lead to a performance decline compared to a more abstract CoT interpretation (understanding the stance from a semantic level). This occurs because overly specific QAPs cause ChatGPT to focus more on the words within a sentence, neglecting higher-level semantic understanding. 
The difference between word-level and semantic-level prompts shows in Figure \ref{Fig222}.
This finding aligns with earlier research observations, as the viewpoints supporting stance predictions in stance detection tasks often do not explicitly appear within the sentence.

In summary, adaptively selecting an appropriate number and type of QAQs, focusing on the data targets or content itself, constitutes a crucial research issue in stance detection tasks.

(3) \textbf{Fine-grained task definition}.
Third, we found that user comments can be challenging to categorize directly into one attitude polarity for some complex targets. For instance, we discovered in our experiments that targets can be framed from different aspects, such as Trump being considered an actor, politician, or businessman. Consequently, user comments may express stances from various perspectives, making it difficult to form a unified conclusion about the attitude to the given target. For example, when given the input "he is a good actor," the stance from an actor's viewpoint should be positive, whereas from a politician's perspective, it may be negative. Therefore, we will further explore the definition of fine-grained stance detection tasks.

\section{Conclusion}
This study investigates the chain-of-thought (CoT) prompting strategy with ChatGPT (GPT-3.5) for stance detection tasks. CoT uses templates as prompts to effectively utilize the model. Our experiments show that CoT achieves state-of-the-art or comparable performance on popular datasets without training, while also discussing limitations when using VLPLMs.

\bibliography{anthology,custom}
\bibliographystyle{acl_natbib}




\end{document}